\documentclass{article} 
\usepackage{iclr2023_conference_supp,times}

\usepackage{amsmath,amsfonts}
\def\bm#1{\mathbf{#1}}

\def\eqref#1{equation~\ref{#1}}

\def\1{\bm{1}}

\DeclareMathAlphabet{\mathsfit}{\encodingdefault}{\sfdefault}{m}{sl}
\SetMathAlphabet{\mathsfit}{bold}{\encodingdefault}{\sfdefault}{bx}{n}

\usepackage{url}
\usepackage{enumitem}
\usepackage{tikz}
\usepackage{comment}
\usepackage{amsmath,amssymb} 
\usepackage{color}
\usepackage{graphicx}
\usepackage{booktabs}
\usepackage{textcomp}  
\usepackage{amsfonts}
\usepackage{multirow}
\usepackage{algorithm}
\usepackage{algorithmic}
\usepackage{float}
\usepackage{tabularx}
\usepackage{array}
\usepackage[misc]{ifsym}
\usepackage{subcaption}
\usepackage{nicefrac}       
\usepackage{wrapfig}
\usepackage{makecell} 

\def\cite{\citep}

\newcommand*{\ie}{\textit{i.e.}}

\usepackage{xcolor}
\usepackage{soul}
\definecolor{my_blue}{HTML}{2b90d9}
\definecolor{my_red}{HTML}{ff5f2e}
\definecolor{my_yellow}{HTML}{FFF2CC}
\definecolor{my_purple}{HTML}{FF8BFF}
\definecolor{my_green}{HTML}{193718}

\usepackage[pagebackref=true,breaklinks=true,colorlinks,bookmarks=false,urlcolor=my_purple]{hyperref}

\title{Joint 2D-3D Multi-Task Learning on Cityscapes-3D: 3D Detection, Segmentation, and Depth Estimation}

\iclrfinalcopy

\author{Hanrong Ye and Dan Xu\\
Department of Computer Science and Engineering \\
The Hong Kong University of Science and Technology (HKUST)\\
Clear Water Bay, Kowloon, Hong Kong \\
\texttt{\{hyeae,danxu\}@cse.ust.hk}
}

\begin{document}

\maketitle

\begin{figure}[h]
    \centering
   \vspace{-10pt} \includegraphics[width=1.\linewidth]{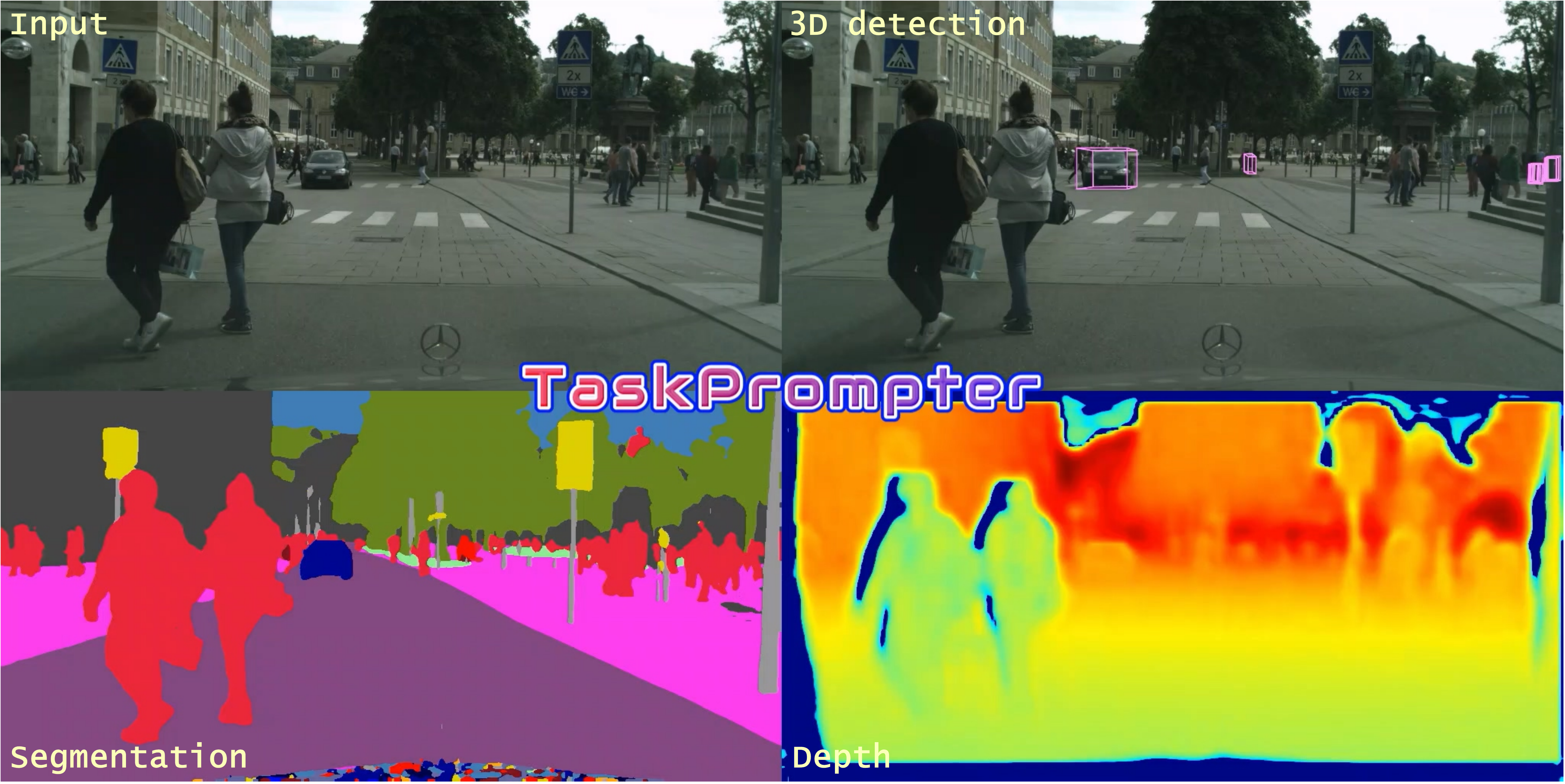}
    \caption{TaskPrompter, trained on Cityscapes-3D, is able to simultaneously generate predictions for various 2D and 3D scene understanding tasks, such as monocular 3D vehicle detection, semantic segmentation, and  monocular depth estimation (\href{https://youtu.be/-eAvl8CLV1g}{video link}).}
    \label{fig:teaser.pdf}
\end{figure}

\begin{abstract}
This report serves as a supplementary document for TaskPrompter~\cite{taskprompter2023}, detailing its implementation on a new joint 2D-3D multi-task learning benchmark based on Cityscapes-3D. TaskPrompter presents an innovative multi-task prompting framework that unifies the learning of (i) task-generic representations, (ii) task-specific representations, and (iii) cross-task interactions, as opposed to previous approaches that separate these learning objectives into different network modules. This unified approach not only reduces the need for meticulous empirical structure design but also significantly enhances the multi-task network's representation learning capability, as the entire model capacity is devoted to optimizing the three objectives simultaneously.
TaskPrompter introduces a new multi-task benchmark based on Cityscapes-3D dataset, which requires the multi-task model to concurrently generate predictions for monocular 3D vehicle detection, semantic segmentation, and monocular depth estimation. These tasks are essential for achieving a joint 2D-3D understanding of visual scenes, particularly in the development of autonomous driving systems. On this challenging benchmark, our multi-task model demonstrates strong performance compared to single-task state-of-the-art methods and establishes new state-of-the-art results on the challenging 3D detection and depth estimation tasks.
 Codes are publicly available \href{https://github.com/prismformore/Multi-Task-Transformer/tree/main/TaskPrompter}{here}.
\end{abstract}

\section{Introduction}
Understanding the pixel-level information of  visual scenes is an important problem in computer vision, with real-world applications in AR/VR, mobile devices, autonomous driving, and so on.
This problem involves the study of many challenging perception tasks, including semantic segmentation, monocular depth estimation, object detection, etc. These tasks share lots of commonalities and can be jointly learned in a unified framework in a multi-task learning manner~\cite{mtlsurvey}.

Multi-task learning has two-fold strengths compared with single-task learning: (i) First, jointly learning multiple tasks in one model is naturally more efficient than learning tasks with different models separately, since different tasks can share some common information captured by task-shared network modules.
(ii) Secondly, as different tasks can help each other via information complementation, they can outperform single-task models with a proper design of multi-task models. Notably, one latest  example of the multi-task model surpassing single-task counterparts is OpenAI's ChatGPT~\cite{brown2020language}.
Therefore, there is a growing interest among researchers in multi-task learning for visual scene perception and understanding.

As one of the latest achievements in multi-task learning,
TaskPrompter~\cite{taskprompter2023} aims to integrate the acquisition of three key perspectives of information in multi-task learning - namely, task-generic representations, task-specific representations, and cross-task interactions - into a single module.
Different from previous encoder-focused~\cite{bachmann2022multimae,xu2022mtformer,bhattacharjee2022mult,xu2022multi,crossstitch,nddr,zhang2021automtl,gao2020mtl,bruggemann2020automated} and decoder-focused~\cite{invpt2022,mti,atrc,papnet,padnet,zhang2021transfer,PSD} methods that separate the learning of these three perspectives of information, TaskPrompter minimizes the necessity for manual designing or intensive searching of modules tailored to different objectives. Instead, it allocates the majority of the model parameters to simultaneously learn the three kinds of information.
To achieve this, TaskPrompter designs a multi-task prompting framework for dense prediction tasks. The key concept is the creation of task-specific learnable tokens, known as ``spatial-channel task prompts", which enable the learning of spatial- and channel-wise task-specific information for each task. These prompts are embedded with task-generic patch tokens from the input image, and together they form the input for a transformer with a specially designed ``Spatial-Channel Task Prompt Learning'' module. The task prompts and patch tokens interact through attention mechanisms in each transformer layer, allowing for the simultaneous learning of task-generic and task-specific representations as well as cross-task interaction. This approach eliminates the need for designing different types of network modules.
To generate prediction maps for multiple dense prediction tasks, TaskPrompter proposes a ``Dense Spatial-Channel Task Prompt Decoding" method. \textbf{For further introduction regarding TaskPrompter, we recommend referring to the published \href{https://openreview.net/pdf?id=-CwPopPJda}{paper}.}

TaskPrompter achieves state-of-the-art performances on several multi-task dense prediction benchmarks, such as the widely-used PASCAL-Context~\cite{everingham2010pascal} and NYUD-v2~\cite{silberman2012indoor}, while also requiring less computational costs compared to previous transformer-based methods.
TaskPrompter can be easily adapted to more diverse multi-task learning problems. To demonstrate this, in this report,  we present a joint 2D-3D multi-task learning benchmark called ``MTCityscapes-3D", which is based on Cityscapes-3D~\cite{gahlert2020cityscapes3d,Cordts2016Cityscapes}. In MTCityscapes-3D, the model is required to simultaneously learn three challenging perception tasks: monocular 3D vehicle detection, semantic segmentation, and monocular depth estimation.
These tasks play a crucial role in the development of an autonomous driving system.
TaskPrompter demonstrates strong performance on MTCityscapes-3D, surpassing the state-of-the-art methods on monocular 3D detection and depth estimation on Cityscapes, while showing competitive performance on semantic segmentation.
The primary focus of this report will be on presenting MTCityscapes-3D, implementing TaskPrompter for this benchmark, and discussing the experimental outcomes.

\section{Joint 2D-3D Multi-Task Learning for Scene Understanding}
\subsection{Multi-Task Cityscapes-3D Benchmark}
To further evaluate TaskPrompter's performance in multi-task scene understanding, we design a unified 2D-3D multi-task scene understanding problem that encompasses the three aforementioned challenging tasks, \ie, monocular 3D vehicle detection (3Ddet), semantic segmentation (Semseg), and monocular depth estimation (Depth), using the Cityscapes-3D dataset~\cite{gahlert2020cityscapes3d}.
We term this benchmark as ``\textbf{Multi-Task Cityscapes-3D Benchmark (MTCityscapes-3D})''.
MTCityscapes-3D comprises 2,975 training images and 500 validation images with fine annotations from Cityscapes-3D. The original image resolution is $1024\times2048$.  For semantic segmentation evaluation, we employ the more challenging 19-class labels. The models are assessed on the validation set for all tasks.
The 3D vehicle detection task~(3Ddet) utilizes the mean detection score (mDS) as its metric, using the official evaluation script provided by Cityscapes-3D.
Importantly, during the experiments, we compare TaskPrompter, which can generate predictions for these tasks concurrently, with the state-of-the-art single-task approaches.

\subsection{Significance of Cityscapes-3D as the Dataset Choice}
Cityscapes~\cite{Cordts2016Cityscapes} is a popular dataset for dense scene understanding, offering labels for tasks including semantic segmentation and depth estimation. Gahlert et al.~\cite{gahlert2020cityscapes3d} further annotated accurate 3D bounding boxes for vehicle detection, leading to the creation of ``Cityscapes-3D".
In contrast to earlier 3D object detection datasets such as nuScenes~\cite{caesar2020nuscenes}, Waymo~\cite{sun2020scalability}, and KITTI~\cite{geiger2013vision}, Cityscapes-3D is specifically tailored for \textit{monocular} 3D object detection. This is due to the fact that the ground-truth depth labels of Cityscapes-3D are derived from stereo cameras, instead of lidar, which frequently leads to discrepancies between images and depth maps~\cite{gahlert2020cityscapes3d}.
Additionally, Cityscapes-3D provides labels for both pitch and roll angles of vehicles, while most previous datasets only focus on yaw. These angles are crucial when roads are not level, a frequent occurrence. Moreover, Cityscapes-3D introduces a more appropriate metric called ``mean detection score (mDS)'' for assessing monocular 3D object detection, better capturing the actual performance.

\subsection{Implementation of TaskPrompter on MTCityscapes-3D}

\subsubsection{Main Network Structure}
Given that Cityscapes-3D images possess a larger spatial resolution and 3Ddet is sensitive to object sizes, we construct TaskPrompter based on the Swin-Base model~\cite{swin}, which retains high-resolution features throughout the network layers. Importantly, since the transformer block in Swin-Base employs local window attention rather than the global attention used in ViT~\cite{vit}, we duplicate the spatial task prompts and integrate them into each local window to facilitate interaction with all patch tokens. Following the computation of window attention, the spatial task prompts from different windows are combined into a single prompt by averaging.
Additionally, at each Patch Merging layer of Swin-Base, the channel number of task prompts is doubled to maintain consistency with the patch token channel numbers.

In the decoding phase, we stitch the Spatial-Task-Prompt Affinity tensors from all windows to form a global affinity tensor, which is then used to query the patch tokens. The learning and decoding of channel task prompts remain unaffected by the Swin-Base architecture and can be executed in the same manner as with the ViT backbone version. We don't use cross-task reweighting in Swin backbone because it is not straightforward to obtain the prompt-to-prompt affinity maps from the window attention.

\subsubsection{Prediction Heads}
For all three tasks, we employ task-specific Conv($3\times3$)-BN-ReLU blocks as prediction heads to produce the final task features, and a linear layer is used to generate the final predictions for Semseg and Depth. Regarding 3Ddet, we utilize the final prediction heads of FCOS-3D~\cite{wang2021fcos3d} to estimate location coordinates, rotation angles, dimensions, object classes, center-ness, and direction classification. For more details about the 3D detection prediction head, please refer to FCOS-3D.

\subsubsection{Optimization}
To reduce computation cost, we decrease the resolution of input images from 1024$\times$2048 to 768$\times$1536. We set the batch size to 2 and train the model for 40k iterations. Notably, the evaluation is still computed at the original resolution.

We use the Adam optimizer with a learning rate of $2\times10^{-5}$ and no weight decay. For the 3D detection task, we apply Non-maximum Suppression with a threshold of 0.3.
The loss weight assigned for semantic segmentation is 100, whereas for depth estimation and 3D detection, it is set to 1.
For 3D detection, in a manner akin to FCOS3D~\cite{wang2021fcos3d}, we employ focal loss~\cite{Lin_2017_ICCV} for object classification, smooth $\mathcal{L}1$ loss for location coordinate and size regression, and cross-entropy loss for direction classification and center-ness regression. For semantic segmentation, we utilize cross-entropy loss, and for depth estimation, we apply $\mathcal{L}1$ loss.

\subsection{Experimental Results}
Table~\ref{tab:sota_cityscapes} presents a performance comparison between state-of-the-art methods (single-task models), the multi-task baseline, and TaskPrompter on the MTCityscapes-3D benchmark.
The multi-task baseline utilizes the same backbone model (\textit{i.e.}, Swin-Base) and identical prediction heads as TaskPrompter, specifically, Conv($3\times3$)-BN-ReLU-Conv($1\times1$) blocks.

We highlight that our work is the first in the literature to concurrently execute all three tasks on this dataset. Our TaskPrompter significantly outperforms the baseline across all tasks, further substantiating the effectiveness of the method. Impressively, TaskPrompter also \textbf{surpasses} state-of-the-art single-task/multi-task models on 3Ddet~\cite{haq2022one} and Depth~\cite{wang2020sdc}, and demonstrates strong performance on highly competitive Semseg when compared to SETR~\cite{SETR} with ViT-B backbone.

\begin{figure}[t]
    \centering
    \includegraphics[width=1.\linewidth]{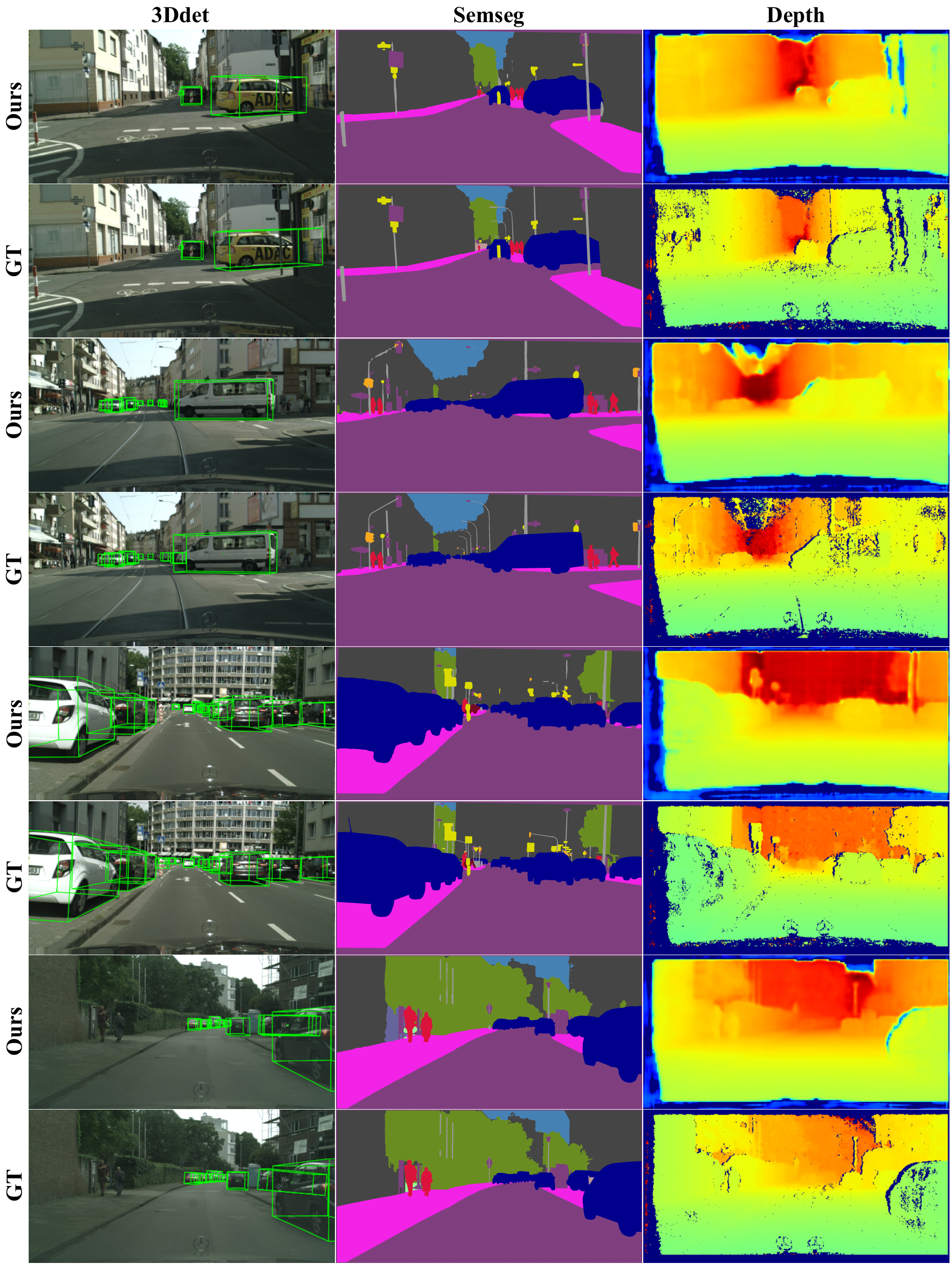}
    \caption{Prediction visualization on MTCityscapes-3D. TaskPrompter can generate competitive results for multiple 2D and 3D scene understanding tasks including monocular 3d vehicle detection, semantic segmentation, and monocular depth estimation simultaneously.}
    \label{fig:qua_cs}
\end{figure}

In Figure~\ref{fig:qua_cs}, we visualize our prediction results on MTCityscapes-3D alongside ground truth labels, revealing that TaskPrompter can concurrently generate competitive results for multiple 2D and 3D scene understanding tasks. These experimental outcomes further suggest that the proposed TaskPrompter can be effectively adapted to other transformer models and task sets.

\begin{table}[t]
\centering
\caption{Performance of 2D/3D scene understanding on Cityscapes-3D. \textit{TaskPrompter achieves better or competitive results against SOTA methods of multiple tasks.} Bold denotes the best. `$\mathbf{\downarrow}$' means lower better and `$\mathbf{\uparrow}$' means higher better.}
\vspace{-10pt}
\label{tab:sota_cityscapes}
\resizebox{0.9\linewidth}{!}{
    \begin{tabular}{c|l|ccccc|cccc}
    \toprule
     \multicolumn{2}{c|}{ \multirow{2}*{ \textbf{Model}} }& \textbf{3Ddet} & \textbf{Semseg}  & \textbf{Depth}\\
    \multicolumn{2}{c|}{}    & mDS $\mathbf{\uparrow}$ & mIoU $\mathbf{\uparrow}$ & RMSE $\mathbf{\downarrow}$  \\
    \midrule
    \multicolumn{1}{c|}{ \multirow{3}*{ \makecell{Single-task\\ Models}}}
    & \textbf{3Ddet}: One-Stage~\cite{haq2022one} & 26.90 & -& -\\
    & \textbf{Semseg}: SETR w/ ViT-B~\cite{SETR} & - & \textbf{78.02}\\
    & \textbf{Depth}: SDC-Depth~\cite{wang2020sdc} & -&-& 6.92\\
    \midrule
    \multicolumn{1}{c|}{ \multirow{2}*{ \makecell{Multi-task\\ Models}}} & Our Baseline & 29.69 & 76.90 & 7.00  \\
      &\multicolumn{1}{l|}{\textbf{TaskPrompter}} & \textbf{32.94} & 77.72  & \textbf{6.78} \\
    \bottomrule
    \end{tabular}}
\end{table}

\section{Conclusion}
In this report, we present a novel multi-task learning benchmark for scene understanding, called MTCityscapes-3D, based on the popular Cityscapes dataset. This benchmark simultaneously evaluates a multi-task model's performance on three tasks: monocular 3D detection, semantic segmentation, and monocular depth estimation, which are crucial components for constructing a comprehensive visual scene understanding system. We assess the performance of the recently proposed TaskPrompter on MTCityscapes-3D and show that TaskPrompter surpasses not only the multi-task baseline but also several state-of-the-art single-task methods. We hope the new multi-task benchmark and the exciting results can inspire further research in multi-task learning for scene understanding, particularly in the context of integrating 2D and 3D tasks crucial for developing autonomous driving systems.

\section*{Acknowledgement}
This technical report is part of the TaskPrompter project~\cite{taskprompter2023} by Hanrong Ye and Dan Xu.
The author would like to thank Tai Wang and Wei-Hong Li for their helpful suggestions.

\clearpage

\bibliography{refers}
\bibliographystyle{iclr2023_conference}

\end{document}